\pdfoutput=1
\documentclass{article}

\usepackage[preprint, nonatbib]{neurips_2024}

\usepackage[utf8]{inputenc} % allow utf-8 input
\usepackage[T1]{fontenc}    % use 8-bit T1 fonts
\usepackage{hyperref}       % hyperlinks
\usepackage{url}            % simple URL typesetting
\usepackage{booktabs}       % professional-quality tables
\usepackage{amsfonts}       % blackboard math symbols
\usepackage{nicefrac}       % compact symbols for 1/2, etc.
\usepackage{microtype}      % microtypography
\usepackage{xcolor}         % colors

\usepackage{colortbl}
\usepackage{graphicx}
\usepackage{multirow} 
\usepackage{amsmath}
\usepackage{mathrsfs}
\usepackage{subcaption}
\usepackage{cleveref}
\usepackage{amsfonts,amssymb}
\usepackage{subcaption}
\usepackage{enumitem}

\title{SparseDrive: End-to-End Autonomous Driving via Sparse Scene Representation}

\author{%
Wenchao Sun\textsuperscript{\textnormal{1,2}} \quad
Xuewu Lin\textsuperscript{\textnormal{2}} \quad
Yining Shi\textsuperscript{\textnormal{1}} \quad
Chuang Zhang\textsuperscript{\textnormal{1}} \quad 
Haoran Wu\textsuperscript{\textnormal{1}} \quad
Sifa Zheng\textsuperscript{\textnormal{1}}
\vspace{0.2cm}\\
$^1$ Tsinghua University \quad
$^2$ Horizon
}
\begin{document}
\maketitle
\begin{abstract}
The well-established modular autonomous driving system is decoupled into different standalone tasks, e.g. perception, prediction and planning, suffering from information loss and error accumulation across modules. In contrast, end-to-end paradigms unify multi-tasks into a fully differentiable framework, allowing for optimization in a planning-oriented spirit. Despite the great potential of end-to-end paradigms, both the performance and efficiency of existing methods are not satisfactory, particularly in terms of planning safety. We attribute this to the computationally expensive BEV (bird's eye view) features and the straightforward design for prediction and planning. To this end, we explore the sparse representation and review the task design for end-to-end autonomous driving, proposing a new paradigm named SparseDrive. Concretely, SparseDrive consists of a symmetric sparse perception module and a parallel motion planner. The sparse perception module unifies detection, tracking and online mapping with a symmetric model architecture, learning a fully sparse representation of the driving scene. For motion prediction and planning, we review the great similarity between these two tasks, leading to a parallel design for motion planner. Based on this parallel design, which models planning as a multi-modal problem, we propose a hierarchical planning selection strategy , which incorporates a collision-aware rescore module, to select a rational and safe trajectory as the final planning output. With such effective designs, SparseDrive surpasses previous state-of-the-arts by a large margin in performance of all tasks, while achieving much higher training and inference efficiency. Code will be avaliable at \url{https://github.com/swc-17/SparseDrive} for facilitating future research.
\end{abstract}
\section{Introduction} \label{introduction}
\begin{figure}[tb]
  \centering
  \begin{subfigure}{0.6\linewidth}
  \begin{subfigure}{1\linewidth}
    \includegraphics[width=1\linewidth]{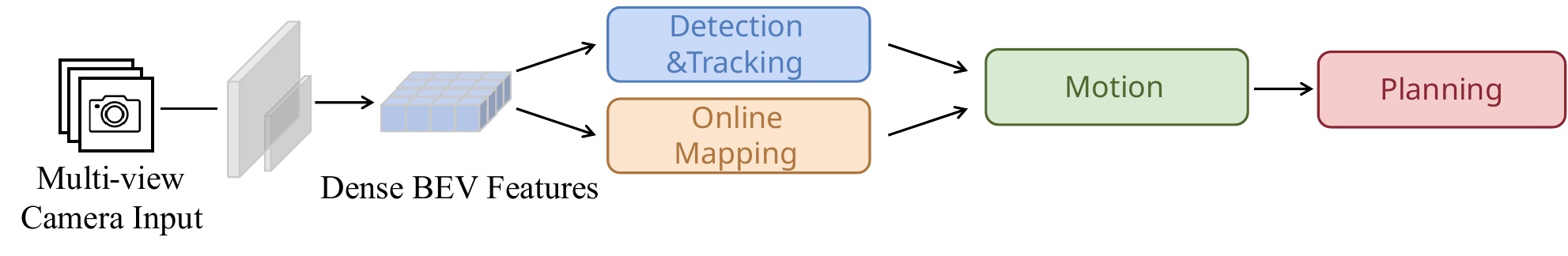}
    \caption{BEV-Centric paradigm.}
    \label{fig:pipeline_bev}
  \end{subfigure}
  \hfill
  \\
  \begin{subfigure}{1\linewidth}
    \includegraphics[width=1\linewidth]{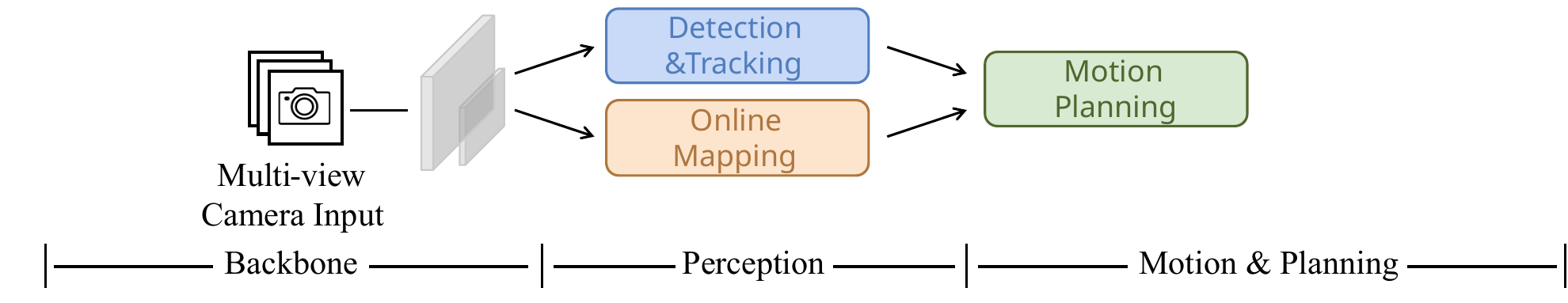}
    \caption{Sparse-Centric paradigm.}
    \label{fig:pipeline_sparse}
  \end{subfigure}
  \label{fig:short}
  \\
  \hfill
  \end{subfigure}
  \hspace{2pt}
  \begin{subfigure}{0.38\linewidth}
  \centering
    \begin{subfigure}{1\linewidth}
    \includegraphics[width=1\linewidth]{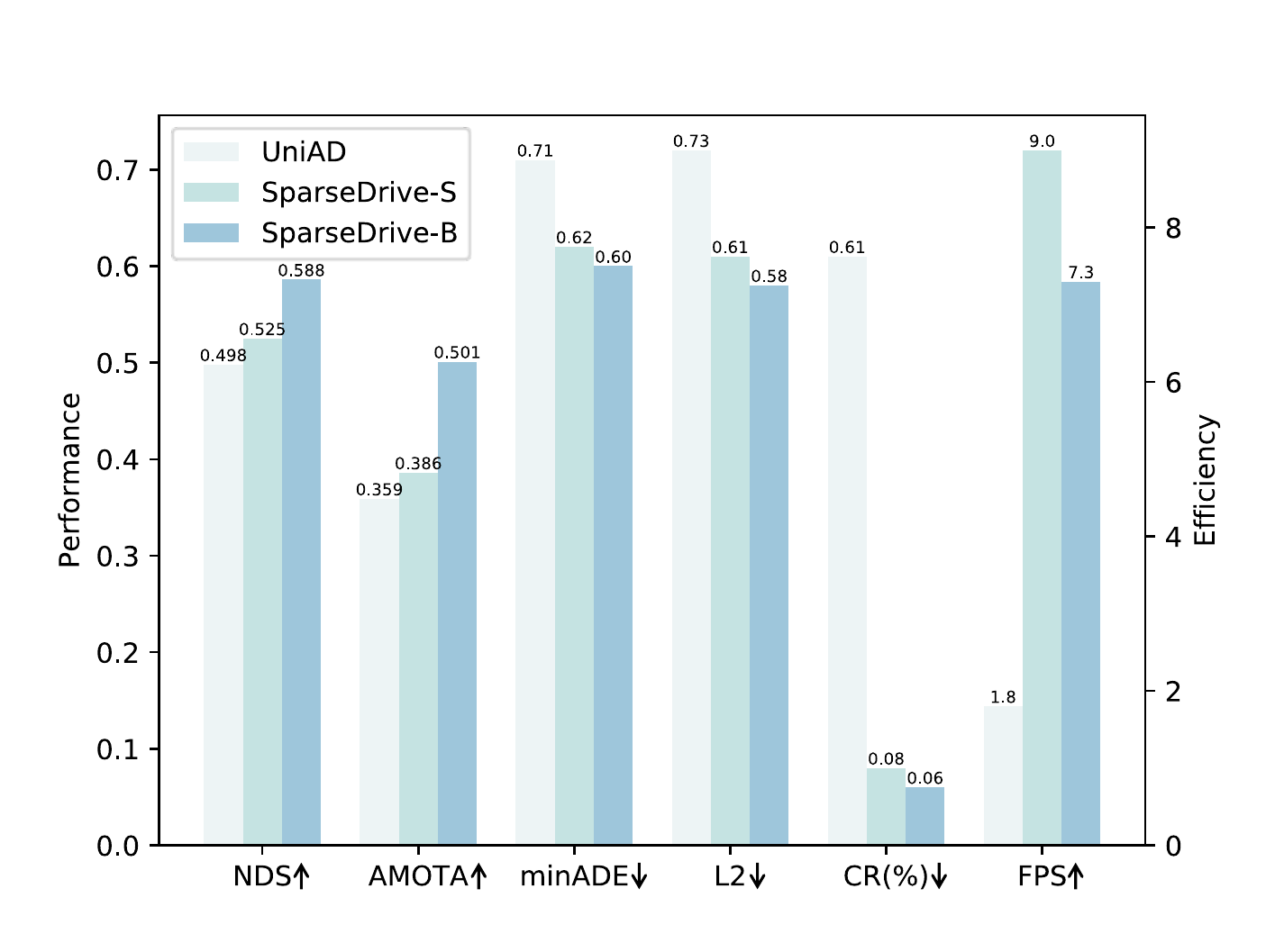}
    \caption{Comparison between our method and privous SOTA\cite{uniad}.}
    \label{fig:comp}
    \end{subfigure}
    \\
    \begin{subfigure}{1\linewidth}
    \end{subfigure}
    \\
    \begin{subfigure}{1\linewidth}
    \end{subfigure}

  \end{subfigure}
  
  \caption{The comparison of various end-to-end paradigms. (a) The BEV-Centric paradigm. (b) The proposed Sparse-Centric paradigm. (c) Performance and efficiency comparison between (a) and (b).}
  \label{fig:pipeline}
\end{figure}

Traditional autonomous driving system is characterized as
modular tasks in sequential order. While advantageous in interpretation and error tracking, it inevitably leads to information loss and accumulative errors across successive modules, thereby limiting the optimal performance potential of the system.

Recently, an end-to-end driving paradigm emerged as a promising research direction. This paradigm integrates all tasks into one holistic model, and can be optimized toward the ultimate pursuit for planning. However, the existing methods\cite{uniad, vad} are not satisfactory in terms of performance and efficiency. On one hand, previous methods rely on computationally expensive BEV features. On the other hand, the straightforward design for prediction and planning limits the model performance. We summarize previous 
methods as BEV-Centric paradigm in Fig. \ref{fig:pipeline_bev}.

To fully leverage the potential of end-to-end paradigm, we review the task design of existing methods, and argue that three main parallels shared between motion prediction and planning are neglected as follows: (1) Aiming at predicting future trajectories of surrounding agents and ego vehicle, both motion prediction and planning should consider the high-order and bidirectional interactions among road agents. However, previous methods typically adopt a sequential design for motion prediction and planning, ignoring the impact of ego vehicle on surrounding agents. (2) Accurate prediction for future trajectories requires semantic information for scene understanding, and geometric information to predict future movement of agents, which is applicable to both motion prediction and planning. While these information are extracted in upstream perception tasks for surrounding agents, it is overlooked for ego vehicle. (3) Both motion prediction and planning are multi-modal problems with inherent uncertainty, but previous methods only predict deterministic trajectory for planning. 

To this end, we propose SparseDrive, a Sparse-Centric paradigm as shown in Fig. \ref{fig:pipeline_sparse}. Specifically, SparseDrive is composed of a symmetric sparse perception module and a parallel motion planner. With the decoupled instance feature and geometric anchor as complete representation of one instance (a dynamic road agent or a static map element), \textbf{Symmetric Sparse Perception} unifies detection, tracking and online mapping tasks with a symmetric model architecture, learning a fully sparse scene representation. In \textbf{Parallel Motion Planner}, a semantic-and-geometric-aware ego instance is first obtained from ego instance initialization module. With the ego instance and surrounding agent instances from sparse perception, motion prediction and planning are conducted simultaneously to get multi-modal trajectories for all road agents. To ensure the rationality and safety for planning, a hierarchical planning selection strategy that incorporating a collision-aware rescore module is applied to select the final planning trajectory from multi-modal trajectory proposals.

With above effective designs, SparseDrive unleashes the great potential of end-to-end autonomous driving, as shown in Fig. \ref{fig:comp}. Without bells and whistles, our base model, SparseDrive-B, greatly reduces the average L2 error by 19.4\% (0.58m vs. 0.72m) and collision rate by 71.4\% (0.06\% vs. 0.21\%). Compared with previous SOTA (state-of-the-art) method UniAD\cite{uniad}, our small model, SparseDrive-S achieves superior performance among all tasks, while running 7.2$\times$ faster for training (20 h vs. 144 h) and 5.0$\times$ faster for inference (9.0 FPS vs. 1.8 FPS).
  
The main contribution of our work are summarized as follows:
\begin{itemize}[leftmargin=*]
\item We explore the sparse scene representation for end-to-end autonomous driving and propose a Sparse-Centric paradigm named SparseDrive, which unifies multiple tasks with sparse instance representation.
\item We revise the great similarity shared between motion prediction and planning, correspondingly leading to a parallel design for motion planner. We further propose a hierarchical planning selection strategy incorporating a collision-aware rescore module to boost the planning performance.
\item On the challenging nuScenes\cite{nuscenes} benchmark, SparseDrive surpasses previous SOTA methods in terms of all metrics, especially the safety-critical metric collision rate, while keeping much higher training and inference efficiency.
\end{itemize}
\section{Related Work}
\subsection{Multi-view 3D Detection}
Multi-view 3D detection is a prerequisite for the safety of autonomous driving system. LSS\cite{lss} utilizes depth estimation to lift image features to 3D space and splats features to BEV plane. Follow-up works apply lift-splat operation to the field of 3D detection, and have made significant improvement in accuracy\cite{bevdet, bevdet4d, bevdepth, bevstereo} and efficiency\cite{bevfusion, bevpoolv2}. Some works\cite{bevformer, bevformerv2, polarformer, gkt} predefine a set of BEV queries and project them to perspective view for feature sampling. Another line of research removes the dependency for dense BEV features. PETR series\cite{petr, petrv2, streampetr} introduce 3D positional encoding and global attention to learn view transformation implicitly. Sparse4D series\cite{sparse4d, sparse4dv2, sparse4dv3} set explicit anchors in 3D space, projecting them to image view to aggregate local features and refine anchors in an iterative fashion.

\subsection{End-to-End Tracking}
Most multi-object tracking (MOT) methods adopt the tracking-by-detection fashion, which relies on post-processing like data association. Such pipeline cannot fully leverage the capabilities of neural networks. Inspired by object queries in \cite{detr}, some works\cite{motr, motrv2, motrv3, trackformer, transtrack, mutr3d} introduce track queries to model the tracked instances in streaming manner. MOTR\cite{motr} proposes tracklet-aware label assignment, which forces the track query to continuously detect the same target and suffers from the conflict between detection and association\cite{motrv2, motrv3}. Sparse4Dv3 demonstrates that the temporally propagated instances already have identity consistency, and achieves SOTA tracking performance with a simple ID assignment process.

\subsection{Online Mapping}
Online mapping is proposed as an alternative of HD map, due to the high cost and vast human efforts in HD map construction. HDMapNet\cite{hdmapnet} groups BEV semantic segmentation with post-processing to get vectorized map instances. VectorMapNet\cite{vectormapnet} utilizes a two-stage auto-regressive transformer for online map construction. MapTR\cite{maptr} models map element as a point set of equivalent permutations, which avoids definition ambiguity of map element. BeMapNet adopts piecewise Bezier curve to describe the details of map elements. StreamMapNet\cite{streammapnet} introduces BEV fusion and query propagation for temporal modeling.

\subsection{End-to-End Motion Prediction}
End-to-end motion prediction is proposed to avoid the cascading error in traditional pipelines. FaF\cite{faf} employs a single convolution network to predict both current and future bounding boxes. IntentNet\cite{intentnet} takes one step further to reason both high level behavior and long term trajectories. PnPNet\cite{pnpnet} introduces an online tracking module to aggregate trajectory level features for motion prediction. ViP3D\cite{vip3d} employs agent queries to perform tracking and prediction, taking images and HD map as input. PIP\cite{pip} replaces human-annotated HD map with local vectorized map.

\subsection{End-to-End Planning}
The research of end-to-end planning has been ongoing since last century\cite{pomerleau1988alvinn}. Early works\cite{codevilla2018end, codevilla2019exploring, prakash2021multi} omit intermediate tasks like perception and motion prediction, which lack interpretability and are difficult to optimize. Some works\cite{stp3, mp3, sadat2020perceive, cui2021lookout} construct explicit cost map from perception or prediction results to enhance interpretability, but rely on hand-crafted rules to select the best trajectory with minimum cost. Recently, UniAD\cite{uniad} proposes a unified query design to integrate various tasks into a goal-oriented model, achieving remarkable performance in perception, prediction and planning. VAD\cite{vad} employs vectorized representation for scene learning and planning constraints. GraphAD\cite{graphad} utilizes graph model for complex interactions in traffic scenes. FusionAD\cite{fusionad} extends end-to-end driving to multi-sensor input. However, previous methods mainly focus on scene learning, and adopt a straightforward design for prediction and planning, without fully considering the similarity between these two tasks, greatly limiting the performance.
\begin{figure}[htbp]
  \centering
  \includegraphics[width=0.8\linewidth]{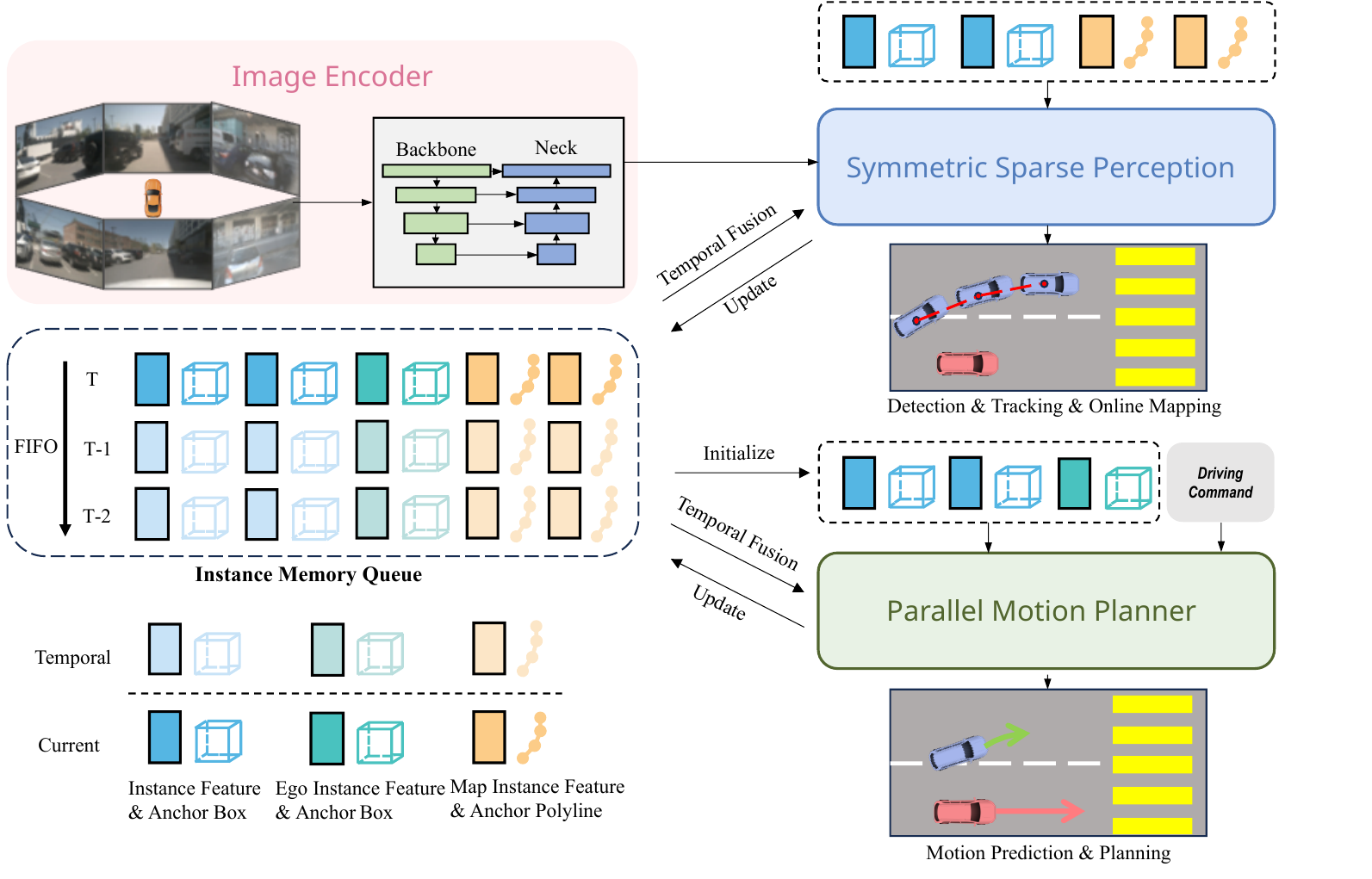}
  \caption{Overview of SparseDrive. SparseDrive first encodes multi-view images into feature maps, then learns sparse scene representation through symmetric sparse perception, and finally perform motion prediction and planning in a parallel manner. An instance memory queue is devised for temporal modeling.}
  \label{fig:overview}
\end{figure}

\section{Method}
\subsection{Overview}
The overall framework of SparseDrive is depicted in Fig. \ref{fig:overview}. Specifically, SparseDrive is consisted of three parts: image encoder, symmetric sparse perception and parallel motion planner. Given multi-view images, the image encoder, including a backbone network and a neck, first encodes images to multi-view multi-scale feature maps $I=\left\{I_s \in \mathbb{R}^{N \times C \times H_s \times W_s} | 1 \leq s \leq S \right\}$, where $S$ is the number of scales and $N$ is the number of camera views. In symmetric sparse perception module, the feature maps $I$ are aggregated into two groups of instances to learn the sparse representation of the driving scene. These two groups of instances, representing surrounding agents and map elements respectively, are fed into parallel motion planner to interact with an initialized ego instance. The motion planner predicts multi-modal trajectories of surrounding agents and ego vehicle simultaneously, and selects a safe trajectory as the final planning result through hierarchical planning selection strategy.

\subsection{Symmetric Sparse Perception}
\begin{figure}[htbp]
  \centering
  \includegraphics[width=0.85\linewidth]{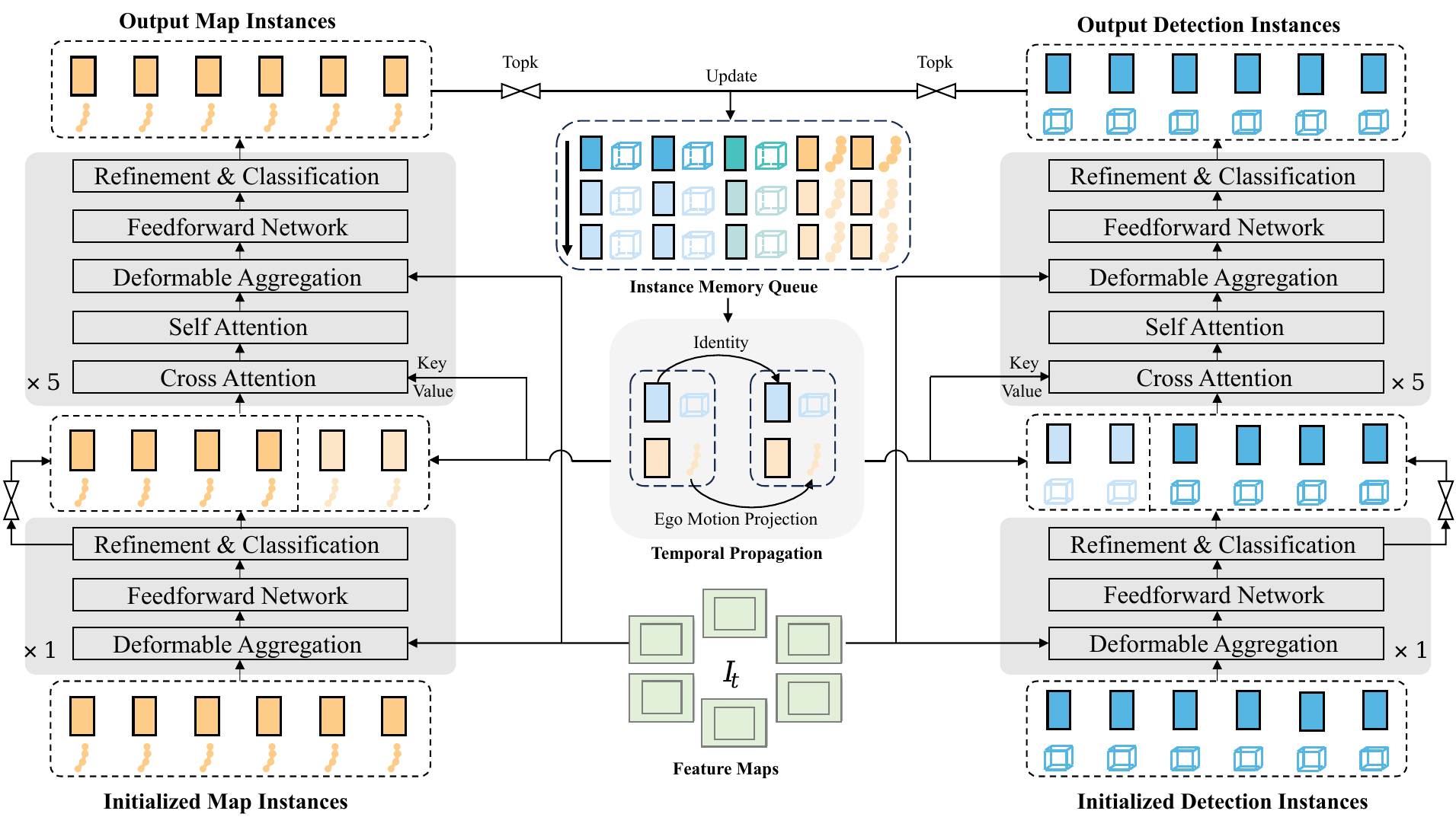}
  \caption{Model architecture of symmetric sparse perception, which unifies detection, tracking and online mapping in a symmetric structure.}
  \label{fig:sparse_perception}
\end{figure}

As shown in Fig. \ref{fig:sparse_perception}, the model structure of sparse perception module exhibits a structural symmetry, unifying detection, tracking and online mapping together.

\paragraph{Sparse Detection.}
Surrounding agents are represented by a group of instance features $F_d \in \mathbb{R}^{N_d \times C}$ and anchor boxes $B_d \in \mathbb{R}^{N_d \times 11}$, where $N_d$ is the number of anchors and $C$ is the feature channel dimension. Each anchor box is formatted  with location, dimension, yaw angle and velocity:
% \[ \left\{x,y,z,\mathrm{ln }w, \mathrm{ln}h, \mathrm{ln}l, \mathrm{sin}yaw, \mathrm{cos}yaw, vx, vy, vz\right\}. \]
\[ \left\{ x, y, z, \ln w, \ln h, \ln l, \sin{yaw}, \cos{yaw}, vx, vy, vz\right\}. \]

The sparse detection branch consists of $N_{dec}$ decoders, including a single non-temporal decoder and $N_{dec}-1$ temporal decoders. Each decoder takes feature maps $I$, instance features $F_d$ and anchor boxes $B_d$ as input, outputs updated instance features and refined anchor boxes. The non-temporal decoder takes randomly initialized instance as input, while the input for temporal decoder come from both current frame and historical frame. Specifically, the non-temporal decoder includes three sub-modules: deformable aggregation, feedforward network (FFN) and the output layer for refinement and classification. The deformable aggregation module generates fixed or learnable keypoints around the anchor boxes $B_d$ and projects them to feature maps $I$ for feature sampling. The instance features $F_d$ are updated by summation with sampled features, and are responsible for predicting the classification scores and the offsets of anchor boxes in the output layer. The temporal decoders have two additional multi-head attention layers: the temporal cross-attention between temporal instances from last frame and current instances, and the self-attention among current instances. In multi-head attention layer, the anchor boxes are transformed into high-dimensional anchor embedding $E_d \in \mathbb{R}^{N_d \times C}$, and serve as the positional encoding.

\paragraph{Sparse Online Mapping.}
Online mapping branch shares the same model structure with detection branch except different instance definition. For static map element, the anchor is formulated as a polyline with $N_p$ points: \[ \left\{x_{0},y_{0},x_{1},y_{1},...,x_{N_p-1},y_{N_p-1} \right\}. \] 

Then all the map elements can be represented by map instance features $F_m \in \mathbb{R}^{N_m \times C}$ and anchor polylines $L_m \in \mathbb{R}^{N_m \times N_p \times 2}$, where $N_m$ is the number of anchor polylines.

\paragraph{Sparse Tracking.}
For tracking, we follow the ID assignment process of Sparse4Dv3\cite{sparse4dv3}: once the detection confidence of an instance surpasses a threshold $T_{thresh}$, it is locked onto a target and assigned with an ID, which remains unchanged throughout temporal propagation. This tracking strategy does not need any tracking constraints, resulting in an elegant and simple symmetric design for sparse perception module.

\subsection{Parallel Motion Planner}
\begin{figure}[htbp]
  \centering
  \includegraphics[width=0.83\linewidth]{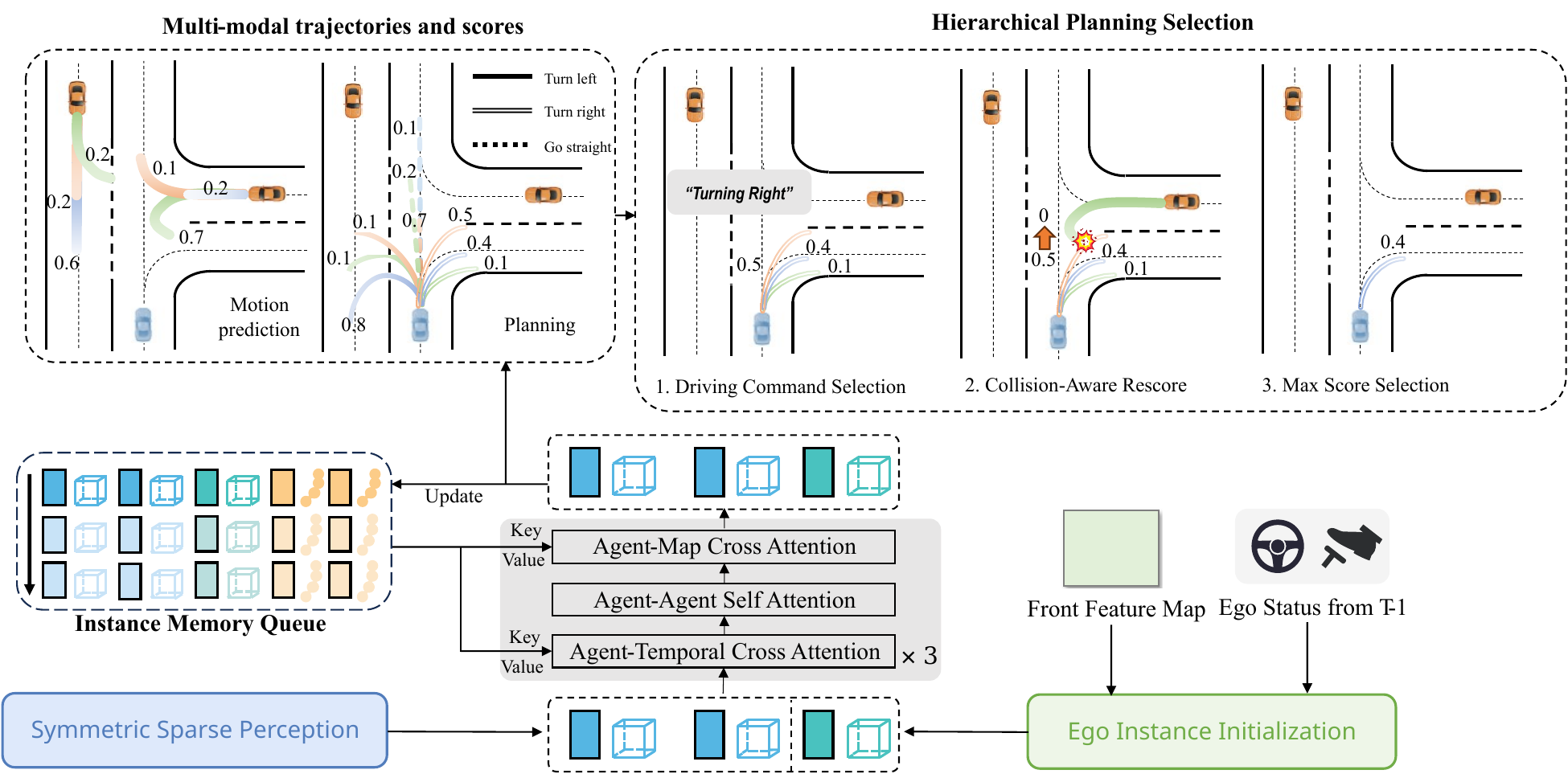}
  \caption{Model structure of parallel motion planner, which performs motion prediction and planning simultaneously and outputs safe planning trajectory.}
  \label{fig:motion_planner}
\end{figure}

As shown in Fig. \ref{fig:motion_planner}, the parallel motion planner consists of three parts: ego instance initialization, spatial-temporal interactions and hierarchical planning selection.

\paragraph{Ego Instance Initialization.}
Similar to surrounding agents, ego vehicle is represented by ego instance feature $F_e \in \mathbb{R}^{1 \times C}$  and ego anchor box $B_e \in \mathbb{R}^{1 \times 11}$. While ego feature is typically randomly initialized in previous methods, we argue that the ego feature also requires rich semantic and geometric information for planning, similar to motion prediction. However, the instance features of surrounding agents are aggregated from image feature maps $I$, which is not feasible for ego vehicle, since ego vehicle is in blind area of cameras. Thus we use the smallest feature map of front camera to initialize the ego instance feature: 
\begin{equation}
F_e = {\rm AveragePool}(I_{front,S})
\end{equation}
There are two advantages in doing so: the smallest feature map has already encoded the semantic context of the driving scene, and the dense feature map serves as a complementary for sparse scene representation, in case there are some blacklist obstacles, which can not be detected in sparse perception.

For ego anchor $B_e$, the location, dimension and yaw angle can be naturally set, as we are aware of these information of ego vehicle. For velocity, directly initialized from ground truth velocity leads to ego status leakage, as illustrated in \cite{ego}. So we add an auxiliary task to decode current ego status $ES_T$, including velocity, acceleration, angular velocity and steering angle. At each frame, we use the predicted velocity from last frame as the initialization of ego anchor velocity.

\paragraph{Spatial-Temporal Interactions.}
To consider the high-level interaction between all road agents, we concatenate the ego instance with surrounding agents to get agent-level instances:
\begin{equation}
F_a={\rm Concat}(F_d, F_e), B_a={\rm Concat}(B_d, B_e)
\end{equation}
As the ego instance is initialized without temporal cues, which is important for planning, we devise an instance memory queue with the size of $(N_d+1) \times H$ for temporal modeling, $H$ is the number of stored frames. Then three types of interactions are performed to aggregate spatial-temporal context: agent-temporal cross-attention, agent-agent self-attention and agent-map cross-attention. Note that in temporal cross-attention of sparse perception module, the instances of current frame interact with all temporal instances, which we name as scene-level interaction. While for agent-temporal cross-attention here, we adopt instance-level interaction to make each instance focus on history information of itself.

Then, we predict the multi-modal trajectories $\tau_m \in \mathbb{R}^{N_{d} \times \mathcal{K}_m \times T_{m} \times 2}$, $\tau_p \in \mathbb{R}^{N_{c} \times \mathcal{K}_p \times T_{p} \times 2 }$ and scores $s_m \in \mathbb{R}^{N_{d} \times \mathcal{K}_m}$, $s_p \in \mathbb{R}^{N_{cmd} \times \mathcal{K}_p}$ for both surrounding agents and ego vehicle, $\mathcal{K}_m$ and $\mathcal{K}_p$ are the number of modes for motion prediction and planning, $T_m$ and $T_p$ are the number of future timestamps for motion prediction and planning, and $N_{cmd}$ is the number of driving command for planning. Following the common practice\cite{uniad, vad}, we use three kinds of driving commands: turn left, turn right and go straight. For planning, we additionally predict current ego status from ego instance feature.

\paragraph{Hierarchical Planning Selection.}
Now we have the multi-modal planning trajectory proposals, to select one safe trajectory $\tau_p^* $ to follow, we design a hierarchical planning selection strategy. First, we select a subset of trajectory proposals $\tau_{p,cmd} \in \mathcal{K}_p \times T_{p} \times 2$, corresponding to the high-level command $cmd$. Then, a novel collision-aware rescore module is adopted to ensure safety. With the motion prediction results, we can assess the collision risk of each planning trajectory proposal, for the trajectory with high collision probability, we reduce the score of this trajectory. In practice, we simply set the score of collided trajectory to $0$. Finally, we select the trajectory with the highest score as the final planning output.

\subsection{End-to-End Learning}
\paragraph{Multi-stage Training.}
The training of SparseDrive is divided into two stages. In stage-1, we train symmetric sparse perception module from scratch to learn the sparse scene representation. In stage-2, sparse perception module and parallel motion planner are trained together with no model weights frozen, fully enjoying the benefit of end-to-end optimization. More training details are provided in Appendix \ref{app:training}.

\paragraph{Loss Functions.}
The loss functions include the loss of four tasks, and the loss of each task can be further divider into classification loss and regression loss. For multi-modal motion prediction and planning task, we adopt the winner-takes-all strategy. For planning, there is an additional regression loss for ego status. We also introduce depth estimation as an auxiliary task to enhance the training stability of the perception module. The overall loss function for end-to-end training is: 
\begin{equation}
\mathcal{L}=\mathcal{L}_{det}+\mathcal{L}_{map}+\mathcal{L}_{motion}+\mathcal{L}_{plan}+\mathcal{L}_{depth}.
\end{equation}
More details about loss functions are provided in Appendix \ref{app:loss_func}.

\section{Experiments} \label{experiments}
Our experiments are conducted on challenging nuScenes\cite{nuscenes} dataset, which contains 1000 complex driving scenes, and each lasts for about 20 seconds. Evaluation metrics of each task are described in Appendix \ref{app:metric}. We have two variants of the model, which only differ in backbone network and input image resolution. For our small model SparseDrive-S, we use ResNet50\cite{resnet} as backbone network and the input image size is 256$\times$704. For our base model, SparseDrive-B, we change backbone network to ResNet101 and input image size to 512$\times$1408. All experiments are conducted on 8 NVIDIA RTX 4090 24GB GPUs. More configuration details are provided in Appendix \ref{app:imp_detail}.

\subsection{Main Results}
We compare with prior state-of-the-arts, both modularized and end-to-end methods. Among end-to-end methods, our lightweight model SparseDrive-S has surpassed previous SOTAs in all tasks, while our base model SparseDrive-B pushes the performance boundaries one step further. The main metrics for each task are marked in grey background in Tables.

\paragraph{Perception.} For 3D detection in Tab. \ref{tab:detection}, SparseDrive achieves \textbf{49.6\%} mAP and \textbf{58.8\%} NDS, yielding a significant improvement of \textbf{+11.6\%} mAP and \textbf{+9.0\%} NDS compared to UniAD\cite{uniad}. For multi-object tracking in Tab. \ref{tab:tracking}, SparseDrive achieves \textbf{50.1\%} AMOTA, and lowest ID switch of \textbf{632}, which surpasses UniAD\cite{uniad} by \textbf{+14.2\%} in terms of AMOTA and gets a \textbf{30.2\%} reduction for ID switch, showing the temporal consistency of tracking tracklet. For online mapping in Tab. \ref{tab:online_mapping}, SparseDrive gets a mAP of \textbf{56.2\%}, also surpassing previous end-to-end method VAD\cite{vad} by \textbf{+8.6\%}.

\begin{table}[t]
\caption{Perception results on nuScenes val dataset. SparseDrive achieves best perfomance on all perception tasks among end-to-end methods. $\dagger$: Reproduced with offcial checkpoint.}
\vspace{5pt}
\begin{subtable}[h]{1.0\textwidth}
\centering
% \resizebox{0.95\linewidth}{!}
\scriptsize
{
\begin{tabular}{l|c|c|ccccc|c}
\toprule
Method & Backbone & mAP$\uparrow$ & mATE$\downarrow$ & mASE$\downarrow$ & mAOE$\downarrow$ & mAVE$\downarrow$ & mAAE$\downarrow$ & \cellcolor{gray!30}NDS$\uparrow$ \\
\midrule
UniAD$^\dagger$~\cite{uniad} & ResNet101 & 0.380 & 0.684 & 0.277 & 0.383 & 0.381 & 0.192 & \cellcolor{gray!30}0.498 \\
SparseDrive-S & ResNet50 & 0.418 & 0.566 & 0.275 & 0.552 & 0.261 & 0.190 & \cellcolor{gray!30}0.525 \\
SparseDrive-B & ResNet101 & \textbf{0.496} & \textbf{0.543} & \textbf{0.269} & \textbf{0.376} & \textbf{0.229} & \textbf{0.179} & \cellcolor{gray!30}\textbf{0.588} \\
\bottomrule
\end{tabular}
}
\caption{3D detection results.}
\label{tab:detection}
\end{subtable}
\begin{subtable}[h]{1.0\textwidth}
\begin{subtable}[h]{0.45\textwidth}
\centering
% \resizebox{0.90\linewidth}{!}
\scriptsize
{
\setlength{\tabcolsep}{0.8mm}
\begin{tabular}{l|cccc}
\toprule
Method & \cellcolor{gray!30}AMOTA$\uparrow$ & AMOTP$\downarrow$ & Recall$\uparrow$ & IDS$\downarrow$ \\
\midrule
ViP3D~\cite{vip3d} & \cellcolor{gray!30}0.217 & 1.625 & 0.363 & - \\
QD3DT~\cite{qd3dt} & \cellcolor{gray!30}0.242 & 1.518 & 0.399 & - \\
MUTR3D~\cite{mutr3d} & \cellcolor{gray!30}0.294 & 1.498 & 0.427 & 3822 \\
\midrule
UniAD~\cite{uniad} & \cellcolor{gray!30}0.359 & 1.320 & 0.467 & 906 \\
SparseDrive-S & \cellcolor{gray!30}0.386 & 1.254 & 0.499 & 886 \\
SparseDrive-B & \cellcolor{gray!30}\textbf{0.501} & \textbf{1.085} & \textbf{0.601} & \textbf{632} \\
\bottomrule
\end{tabular}
}
\caption{Multi-object tracking results.}
\label{tab:tracking}
\end{subtable}
\hfill
\begin{subtable}[h]{0.55\textwidth}
% \resizebox{0.92\linewidth}{!}
\scriptsize
{
\setlength{\tabcolsep}{0.6mm}
\begin{tabular}{l|ccc|c}
\toprule
Method & $AP_{ped}\uparrow$ & $AP_{divider}\uparrow$ & $AP_{boundry}\uparrow$ & \cellcolor{gray!30}mAP$\uparrow$ \\
\midrule
HDMapNet~\cite{hdmapnet} & 14.4 & 21.7 & 33.0 & \cellcolor{gray!30}23.0 \\
VectorMapNet~\cite{vectormapnet} & 36.1 & 47.3 & 39.3 & \cellcolor{gray!30}40.9 \\
MapTR~\cite{maptr} & \textbf{56.2} & \textbf{59.8} & \textbf{60.1} & \cellcolor{gray!30}\textbf{58.7} \\
\midrule
% VAD-Tiny & 31.7 & 42.2 & 45.6 & 39.8 \\
VAD$^\dagger$~\cite{vad} & 40.6 & 51.5 & 50.6 & \cellcolor{gray!30}47.6 \\
SparseDrive-S & 49.9 & \textbf{57.0} & 58.4 & \cellcolor{gray!30}55.1 \\
SparseDrive-B & \textbf{53.2} & 56.3 & \textbf{59.1} & \cellcolor{gray!30}\textbf{56.2} \\
\bottomrule
\end{tabular}
}
\caption{Online mapping results.}
\label{tab:online_mapping}
\end{subtable}
\end{subtable}
\end{table}

\paragraph{Prediction.} For motion prediction in Tab. \ref{tab:motion}, SparseDrive achieves the best performance with \textbf{0.60m} minADE, \textbf{0.96m} minFDE, \textbf{13.2\%} MissRate and \textbf{0.555} EPA. Compared with UniAD\cite{uniad}, SparseDrive reduces errors by \textbf{15.5\%} and \textbf{5.9\%} on minADE and minFDE respectively.

\paragraph{Planning.} For planning in Tab. \ref{tab:planning}, among all methods, SparseDrive achieves a remarkable planning performance, with the lowest L2 error of \textbf{0.58m} and collision rate of \textbf{0.06\%}. Compared with previous SOTA VAD\cite{vad}, SparseDrive reduces L2 error by \textbf{19.4\%} and collision rate by \textbf{71.4\%}, demonstrating the effectiveness and safety of our method. 

\paragraph{Efficiency.} As shown in Tab. \ref{tab:efficiency}, besides the excellent performance, SparseDrive also achieves much higher efficiency for both training and inference. With the same backbone network, our base model achieves \textbf{4.8$\times$} faster in training and \textbf{4.1$\times$} faster in inference, compared with UniAD\cite{uniad}. Our lightweight model can achieve \textbf{7.2 $\times$} and \textbf{5.0$\times$} faster in training and inference. \textbf{}

\begin{table}[t]
\caption{Motion prediction and planning results on nuScenes val dataset. SparseDrive outperforms previous methods by a large margin. $\dagger$: Reproduced with offcial checkpoint. $^*$: LiDAR-based methods.}
\vspace{5pt}
\begin{subtable}[h]{0.4\textwidth}
\centering
\scriptsize
{
\setlength{\tabcolsep}{0.8mm}
\begin{tabular}{l|cccc}
\toprule
Method & \cellcolor{gray!30}minADE($m$)$\downarrow$ & minFDE($m$)$\downarrow$ & MR$\downarrow$ & EPA$\uparrow$  \\
\midrule
Cons Pos.~\cite{uniad} & \cellcolor{gray!30}5.80 & 10.27 & 0.347 & - \\
Cons Vel.~\cite{uniad} & \cellcolor{gray!30}2.13 & 4.01 & 0.318 & - \\
Traditional~\cite{vip3d} & \cellcolor{gray!30}2.06 & 3.02 & 0.277 & 0.209 \\
PnPNet~\cite{pnpnet} & \cellcolor{gray!30}1.15 & 1.95 & 0.226 & 0.222 \\
ViP3D~\cite{vip3d} & \cellcolor{gray!30}2.05 & 2.84 & 0.246 & 0.226 \\
\midrule
UniAD\cite{uniad} & \cellcolor{gray!30}0.71 & 1.02 & 0.151 & 0.456 \\
SparseDrive-S & \cellcolor{gray!30}0.62 & 0.99 & 0.136 & 0.482 \\
SparseDrive-B & \cellcolor{gray!30}\textbf{0.60} & \textbf{0.96} & \textbf{0.132} & \textbf{0.555} \\
\bottomrule
\end{tabular}
}
\caption{Prediction results.}
\label{tab:motion}
\end{subtable}
\hfill
\begin{subtable}[h]{0.6\textwidth}
\centering
\scriptsize
{
\setlength{\tabcolsep}{0.8mm}
\begin{tabular}{l|cccc|cccc}
\toprule
\multirow{2}{*}{Method} &
\multicolumn{4}{c|}{L2($m$)$\downarrow$} & 
\multicolumn{4}{c}{Col. Rate(\%)$\downarrow$} \\
& 1$s$ & 2$s$ & 3$s$ & \cellcolor{gray!30}Avg. & 1$s$ & 2$s$ & 3$s$ & \cellcolor{gray!30}Avg.\\
\midrule
FF$^*$~\cite{ff} & 0.55 & 1.20 & 2.54 & \cellcolor{gray!30}1.43 & 0.06 & 0.17 & 1.07 & \cellcolor{gray!30}0.43 \\
EO$^*$~\cite{eo} & 0.67 & 1.36 & 2.78 & \cellcolor{gray!30}1.60 & 0.04 & 0.09 & 0.88 & \cellcolor{gray!30}0.33 \\
\midrule
ST-P3~\cite{stp3} & 1.33 & 2.11 & 2.90 & \cellcolor{gray!30}2.11 & 0.23 & 0.62 & 1.27 & \cellcolor{gray!30}0.71 \\
UniAD$^\dagger$~\cite{uniad} & 0.45 & 0.70 & 1.04 & \cellcolor{gray!30}0.73 & 0.62 & 0.58 & 0.63  & \cellcolor{gray!30}0.61 \\
VAD$^\dagger$~\cite{vad} & 0.41 & 0.70 & 1.05 & \cellcolor{gray!30}0.72 & 0.03 & 0.19 & 0.43  & \cellcolor{gray!30}0.21 \\ 
SparseDrive-S & \textbf{0.29} & 0.58 & 0.96 & \cellcolor{gray!30}0.61 & \textbf{0.01} & 0.05 & 0.18 & \cellcolor{gray!30}0.08 \\
SparseDrive-B &\textbf{0.29} & \textbf{0.55} & \textbf{0.91} & \cellcolor{gray!30}\textbf{0.58} & \textbf{0.01} & \textbf{0.02} & \textbf{0.13} & \cellcolor{gray!30}\textbf{0.06} \\
\bottomrule
\end{tabular}
}
\caption{Planning results.}
\label{tab:planning}
\end{subtable}
\end{table}

\begin{table}[htbp]
\centering
\caption{Efficiency comparison results. SparseDrive achieves high efficiency for both training and inference. Training time and FPS for UniAD are measured on 8 and 1 NVIDIA Tesla A100 GPUs respectively. Training time and FPS for SparseDrive are measured on 8 and 1 NVIDIA Geforce RTX 4090 GPUs respectively.}
\label{tab:efficiency}
\vspace{5pt}
\resizebox{0.90\linewidth}{!}
{    
\begin{tabular}{c|ccc|cccc}
\toprule
\multirow{2}{*}{Method} &
\multicolumn{3}{c|}{Training Efficiency} & 
\multicolumn{4}{c}{Inference Efficiency} \\
& GPU Memory (G) & Batch Size & Time (h) & GPU Memory (M) & FLOPs (G) & Params (M) & FPS\\
\midrule
UniAD~\cite{uniad} & 50.0 & 1 & 48 + 96 & 2451 & 1709 & 125.0 & 1.8 \\
\midrule
SparseDrive-S & 15.2 & 6 & 18 + 2 & 1294 & 192 & 85.9 & 9.0 \\
SparseDrive-B & 17.6 & 4 & 26 + 4 & 1437 & 787 & 104.7 & 7.3 \\
\bottomrule
\end{tabular}
}
\end{table} 

\subsection{Ablation Study}
We conduct extensive ablation studies to demonstrate the effectiveness of our design choices. We use SparseDrive-S as the default model for ablation experiments.

\paragraph{Effect of designs in Motion Planner.}
To underscore the significance of considering similarity between prediction and planning, we devised several specific experiments, as shown in Tab. \ref{tab:ablation_motion_planner}. ID-2 ignores the impact of ego vehicle on surrounding agents by changing the parallel design for prediction and planning to sequential order, leading to worse performance for motion prediction and collision rate. ID-3 randomly initializes ego instance feature and set all parameters of ego anchor to 0. Removing the semantic and geometric information of ego instance leads to performance degradation in both L2 error and collision rate. ID-4 takes planning as a deterministic problem and only outputs one certain trajectory, resulting in highest collision rate. Moreover, ID-5 removes the instance-level agent-temporal cross-attention, seriously degrading the L2 error to 0.77m. For collision-aware rescore, we have detailed discussion in the following paragraph.

\paragraph{Collision-Aware Rescore.} In previous methods\cite{uniad, graphad}, a post-optimization strategy is adopted to ensure safety based on perception results. However, we argue that this strategy breaks the end-to-end paradigm, resulting in serious degradation in L2 error, as shown in Tab. \ref{tab:ablation_CAR}. Moreover, under our re-implemented collision rate metric, the post-optimization does not make planning safer, but rather more dangerous. By contrast, our collision-aware rescore module reduces collision rate from 0.12\% to 0.08\%, with negligible increase in L2 error, showing the superiority of our method.

\paragraph{Multi-modal planning.}
We conduct experiments on the number of planning modes. As shown in Tab. \ref{tab:ablation_planning_mode}, with the number of planning modes increases, the planning performance improves continuously until saturated at 6 modes, again proving the importance of multi-modal planning.

\begin{table}[htbp]
\centering
\caption{Ablation for designs in parallel motion planner. "PAL" means parallel design for motion prediction and planning task; "EII" means ego instance initialization; "MTM" means multiple mode for planning; "ATA" means agent-temporal cross-attention; "CAR" means collision-aware rescore.}
\label{tab:ablation_motion_planner}
\vspace{5pt}
% \resizebox{0.90\linewidth}{!}
\scriptsize
{    
\setlength{\tabcolsep}{1.5mm}
\begin{tabular}{l|ccccc|ccc|cccc|cccc}
\toprule
\multirow{2}{*}{ID} &
\multirow{2}{*}{PAL} & 
\multirow{2}{*}{EII} & 
\multirow{2}{*}{MTM} &
\multirow{2}{*}{ATA} & 
\multirow{2}{*}{CAR} &
\multicolumn{3}{c|}{Prediction} & 
\multicolumn{4}{c|}{Planning L2($m$)} & 
\multicolumn{4}{c}{Planning Coll.(\%)} \\
&&&&&& \cellcolor{gray!30}minADE & minFDE & MR & 1$s$ & 2$s$ & 3$s$ & \cellcolor{gray!30}Avg. & 1$s$ & 2$s$ & 3$s$ & \cellcolor{gray!30}Avg.\\
\midrule
1 & \checkmark & \checkmark & \checkmark & \checkmark & \checkmark & \cellcolor{gray!30}0.623 & \textbf{0.987} & 0.136 & \textbf{0.29} & \textbf{0.58} & 0.96 & \cellcolor{gray!30}\textbf{0.61} & \textbf{0.01} & \textbf{0.05} & \textbf{0.18} & \cellcolor{gray!30}\textbf{0.08} \\
2 & & \checkmark & \checkmark & \checkmark & \checkmark & \cellcolor{gray!30}0.641 & 1.008 & 0.138 & 0.30 & 0.58 & 0.95 & \cellcolor{gray!30}\textbf{0.61} & 0.02 & 0.06 & 0.23 & \cellcolor{gray!30}0.10\\
3 & \checkmark &  & \checkmark & \checkmark & \checkmark & \cellcolor{gray!30}\textbf{0.621} & 0.988 & \textbf{0.135} & 0.31 & 0.60 & 0.98 & \cellcolor{gray!30}0.63 & 0.03 & 0.07 & 0.21 &  \cellcolor{gray!30}0.11\\
4 & \checkmark & \checkmark & & \checkmark & \checkmark & \cellcolor{gray!30}0.626 & 1.002 & 0.136 & 0.33 & 0.66 & 1.08 &\cellcolor{gray!30}0.69 & 0.03 & 0.11 & 0.60 & \cellcolor{gray!30}0.25 \\
5 & \checkmark & \checkmark & \checkmark & & \checkmark & \cellcolor{gray!30}0.634 & 1.003 & 0.138 & 0.40 & 0.74 & 1.16 & \cellcolor{gray!30}0.77 & 0.02 & 0.13 & 0.32 & \cellcolor{gray!30}0.16 \\
6 & \checkmark & \checkmark & \checkmark & \checkmark &  & \cellcolor{gray!30}0.623 & \textbf{0.987} & 0.136 & \textbf{0.29} & \textbf{0.58} & \textbf{0.95} & \cellcolor{gray!30}\textbf{0.61} & \textbf{0.01} & 0.06 & 0.30 & \cellcolor{gray!30}0.12 \\
\bottomrule
\end{tabular}
}
\end{table} 
\begin{table}[htbp]
\centering
\caption{Ablation for collision-aware rescore and post-optimization in \cite{uniad}.}
\label{tab:ablation_CAR}
\vspace{5pt}
% \resizebox{0.90\linewidth}{!}
\scriptsize
{    
\begin{tabular}{c|cc|cccc|cccc}
\toprule
\multirow{2}{*}{Method} &
\multirow{2}{*}{CAR} &
\multirow{2}{*}{Post-optim.} &
\multicolumn{4}{c|}{Planning L2($m$)} & 
\multicolumn{4}{c}{Planning Coll.(\%)} \\
&&& 1$s$ & 2$s$ & 3$s$ & \cellcolor{gray!30}Avg. & 1$s$ & 2$s$ & 3$s$ & \cellcolor{gray!30}Avg.\\
\midrule
UniAD\cite{uniad} & & & 0.32 & 0.58 & 0.94 & \cellcolor{gray!30}0.61 & 0.15 & 0.24 & 0.36  & \cellcolor{gray!30}0.25 \\
UniAD\cite{uniad} & & \checkmark & 0.45 & 0.70 & 1.04 & \cellcolor{gray!30}0.73 & 0.62 & 0.58 & 0.63  & \cellcolor{gray!30}0.61 \\
\midrule
SparseDrive & & & \textbf{0.29} & \textbf{0.58} & \textbf{0.95} & \cellcolor{gray!30}\textbf{0.61} & \textbf{0.01} & 0.06 & 0.30 & \cellcolor{gray!30}0.12 \\
SparseDrive & \checkmark & & \textbf{0.29} & \textbf{0.58} & 0.96 & \cellcolor{gray!30}\textbf{0.61} & \textbf{0.01} & \textbf{0.05} & \textbf{0.18} & \cellcolor{gray!30}\textbf{0.08} \\
SparseDrive & & \checkmark &  0.44 & 0.73 & 1.11 & \cellcolor{gray!30}0.76 & 0.29 & 0.21 & 0.38 & \cellcolor{gray!30}0.30 \\

\bottomrule
\end{tabular}
}
\end{table} 

\begin{table}[htbp]
\centering
\caption{Ablation for planning mode.}
\label{tab:ablation_planning_mode}
\vspace{5pt}
% \resizebox{0.90\linewidth}{!}
\scriptsize
{    
\begin{tabular}{c|cccc|cccc}
\toprule
\multirow{2}{*}{Number of mode} &
\multicolumn{4}{c|}{Planning L2($m$)} & 
\multicolumn{4}{c}{Planning Coll.(\%)} \\
& 1$s$ & 2$s$ & 3$s$ & \cellcolor{gray!30}Avg. & 1$s$ & 2$s$ & 3$s$ &\cellcolor{gray!30}Avg.\\
\midrule
1 & 0.33 & 0.66 & 1.08 &\cellcolor{gray!30}0.69 & 0.03 & 0.11 & 0.60 & \cellcolor{gray!30}0.25 \\
2 & 0.33 & 0.65 & 1.08 & \cellcolor{gray!30}0.69 & 0.01 & 0.12 & 0.42 & \cellcolor{gray!30}0.18 \\
3 & 0.30 & 0.59 & 0.97 & \cellcolor{gray!30}0.62 & \textbf{0.00} & 0.08 & 0.43 & \cellcolor{gray!30}0.17\\
6 & \textbf{0.29} & \textbf{0.57} & \textbf{0.95} & \cellcolor{gray!30}\textbf{0.61} & 0.01 & \textbf{0.03} & \textbf{0.17} & \cellcolor{gray!30}\textbf{0.07}\\
9 & 0.33 & 0.63 & 1.04 & \cellcolor{gray!30}0.66 & 0.01 & 0.09 & 0.36 & \cellcolor{gray!30}0.15\\
\bottomrule
\end{tabular}
}
\end{table}

% \subsection{Qualitative Results}
% \input{Figures/vis}
% We visualize the results of all tasks in Fig. \ref{fig:vis}. More visualizations are provided in Appendix \ref{app:vis}.

\section{Conclusion and Future Work} \label{conclusion}

\paragraph{Conclusion.} In this work, we explore the sparse scene representation and review the task design in the realm of end-to-end autonomous driving. The resulting end-to-end paradigm SparseDrive achieves both remarkable performance and high efficiency. We hope the impressive performance of SparseDrive can inspire the community to rethink the task design for end-to-end autonomous driving and promote technological progress in this field.

\paragraph{Future work.} There still are some limitations in our work. First, the performance of our end-to-end model still falls behind the single-task method, for example, the online mapping task. Second, the scale of the dataset is not large enough to exploit the full potential of end-to-end autonomous driving, and the open-loop evaluation cannot comprehensively represent the model performance. We leave these problems for future exploration.
\newpage
\small{
\bibliographystyle{plain}
\bibliography{ref}
}
\newpage
\appendix

\section{Metrics} \label{app:metric}
The evaluation for detection and tracking follows standard evaluation protocols\cite{nuscenes}. For detection, we use mean Average Precision(\textbf{mAP}), mean Average Error of
Translation(\textbf{mATE}), Scale(\textbf{mASE}), Orientation(\textbf{mAOE}), Velocity(\textbf{mAVE}), Attribute(\textbf{mAAE}) and nuScenes Detection Score(\textbf{NDS}) to evaluate the model performance. For tracking, we use Average Multi-object Tracking Accuracy(\textbf{AMOTA}), Average Multi-object Tracking Precision(\textbf{AMOTP}), \textbf{RECALL}, and Identity Switches(\textbf{IDS}) as the metrics. For online mapping, we calculate the Average Precision(\textbf{AP}) of three map classes: lane divider, pedestrian crossing and road boundary, then average across all classes to get mean Average Precision(\textbf{mAP}). For motion prediction, we employ metrics including minimum Average Displacement Error(\textbf{minADE}), minimum Final Displacement Error(\textbf{minFDE}), Miss Rate(\textbf{MR}) and End-to-end Prediction Accuracy(\textbf{EPA}) proposed in \cite{vip3d}. The motion prediction benchmark is aligned with UniAD\cite{uniad}.

For planning, we adopt commonly used L2 error and collision rate to evaluate the planning performance. The evaluation of L2 error is aligned with VAD\cite{vad}. For collision rate, there are two drawbacks in previous \cite{uniad, vad} implementation, resulting in inaccurate evaluation in planning performance. On one hand, previous benchmark convert obstacle bounding boxes into occupancy map with a grid size of 0.5m, resulting in false collisions in certain cases, e.g. ego vehicle approaches obstacles that smaller than a single occupancy map pixel\cite{admlp}. (2) The heading of ego vehicle is not considered and assumed to remain unchanged\cite{ego}. To accurately evaluate the planning performance, we account for the changes in ego heading by estimating the yaw angle through trajectory points, and assess the presence of a collision by examining the overlap between the bounding boxes of ego vehicle and obstacles. We reproduce the planning results on our benchmark with official checkpoints\cite{uniad, vad} for a fair comparison.

\section{Implementation Details} \label{app:imp_detail}
\subsection{Perception}
For sparse perception module, we set the number of decoder layer $N_{dec}$ to 6, which are 1 non-temporal decoder and 5 temporal decoders. The location for anchor boxes $B_{d}$ and anchor polylines $L_{m}$ are obtained by K-Means clustering on the training set, and other parameters of anchor boxes are initialized with $ \left\{1,1,1,0,1,0,0,0\right\} $. Each map element is represented by 20 points. The number of anchor boxes $N_d$ and polylines $N_m$ are set to 900 and 100 respectively, and the number of temporal instances for detection and online mapping are 600 and 33. The tracking threshold $T_{thresh}$ is set to 0.2. For detection, the perception range is a circle with a radius of 55m. For online mapping, the perception range is 60m $\times$ 30m longitudinally and laterally. For multi-head attention, we adopt Flash Attention\cite{flashattention} to  save the GPU memory.

\subsection{Motion Planner}
The number of stores frames $H$ in instance memory queue is 3. The number of mode $\mathcal{K}_m$ for motion prediction and $\mathcal{K}_p$ for planning  are both set to 6. The number of future timestamps $\mathcal{T}_m$ for motion prediction and $\mathcal{T}_p$ for planning are set to 12 and 6 respectively.
After spatial-temporal interactions in motion planner, we decode ego status of current frame with ego feature $F_e$ using a multi-layer perceptron (MLP): 
\begin{equation}
ES_T = MLP(F_e)
\end{equation}

For multi-modal trajectories and scores, we use K-Means clustering to obtain the prior intention points and transform them into motion mode queries $MQ_m \in \mathbb{R}^{\mathcal{K}_m \times C} $ and planning mode queries $MQ_p \in \mathbb{R}^{ N_{cmd} \times \mathcal{K}_p \times C} $ with sinusoidal position encoding PE(·), then we add mode queries with agent instance features, decode trajectories and scores with MLPs:
% \[\tau_m = MLP(F_d + MQ_m), s_m = MLP(F_d + MQ_m), \\ \tau_p = MLP(F_e + MQ_p), s_p = MLP(F_e + MQ_p) \].
\begin{gather}
\tau_m = MLP(F_d + MQ_m), \\
s_m = MLP(F_d + MQ_m), \\
\tau_p = MLP(F_e + MQ_p), \\
s_p = MLP(F_e + MQ_p)
\end{gather}
In collision-aware rescore module, we utilize the two most confident trajectories in motion prediction to determine whether ego vehicle will collide with surrounding obstacles.  

\subsection{Loss Functions} \label{app:loss_func}
For perception, the \textit{Hungarian algorithm} is adopted to match each ground truth with one predicted value. The detection loss is a linear combination of a Focal loss\cite{focalloss} for classification and an L1 loss for box regression:
\begin{equation}
L_{det} = \lambda_{det\_cls}L_{det\_cls} + \lambda_{det\_reg}L_{det\_reg}.
\end{equation}
As there are no tracking constraints in ID assignment process, we do not have a track loss. The online mapping loss is similar to detection loss: 
\begin{equation}
L_{map} = \lambda_{map\_cls}L_{map\_cls} + \lambda_{map\_reg}L_{map\_reg}.
\end{equation}
For depth estimation, we use L1 loss for regression:
\begin{equation}
L_{depth} = \lambda_{depth}L_{depth}.
\end{equation}
The loss weights are set as follows: $\lambda_{det\_cls}=2$, $\lambda_{det\_reg}=0.25$, $\lambda_{map\_cls}=1$, $\lambda_{map\_reg}=10$, $\lambda_{depth}$ = 0.2.

For motion prediction and planning, we calculate average displacement error (ADE) between multi-model output and ground truth trajectory, the trajectory with lowest ADE is considered as positive sample and rest are negative samples. For planning, ego status is additionally predicted. We also use Focal loss for classification and L1 loss for regression: 
\begin{align}
L_{motion\_planning} = \lambda_{motion\_cls}L_{motion\_cls} + \lambda_{motion\_reg}L_{motion\_reg} \notag \\
+ \lambda_{plan\_cls}L_{plan\_cls} + \lambda_{plan\_reg}L_{plan\_reg} + \lambda_{plan\_status}L_{plan\_status},
\end{align}
where $\lambda_{motion\_cls}=0.2$, $\lambda_{motion\_reg}=0.2$, $\lambda_{plan\_cls}=0.5$, $\lambda_{plan\_reg}=1.0$, $\lambda_{plan\_status}=1.0$.

\subsection{Training Details} \label{app:training}
We use AdamW optimizer\cite{adamw} and Cosine Annealing\cite{cosine} scheduler for model training. The training hyperparameters are listed in Tab. \ref{tab:training_details}.

\begin{table}[htbp]
\centering
\caption{Training hyperparameters.}
\label{tab:training_details}
\vspace{5pt}
\scalebox{0.8}
{
\begin{tabular}{l|cccccc}
\toprule
Model &  Training stage & Batch Size & Epochs & Lr & Backbone lr scale & Weight decay \\
\midrule
SparseDrive-S & stage-1 & 8 & 100 & $4\times10^{-4}$ & 0.5 & $1\times10^{-3}$ \\
SparseDrive-S & stage-2 & 6 & 10 & $3\times10^{-4}$ & 0.1 & $1\times10^{-3}$ \\
\midrule
SparseDrive-B & stage-1 & 4 & 80 & $3\times10^{-4}$ & 0.1 & $1\times10^{-3}$ \\
SparseDrive-B & stage-2 & 4 & 10 & $3\times10^{-4}$ & 0.1 & $1\times10^{-3}$ \\
\bottomrule
\end{tabular}
}
\vspace{5pt}
\end{table} 

\section{Visualization} \label{app:vis}
\begin{figure}[htbp]
  \centering
  \begin{subfigure}{0.8\linewidth}
  \centering
  \includegraphics[width=1.0\linewidth]{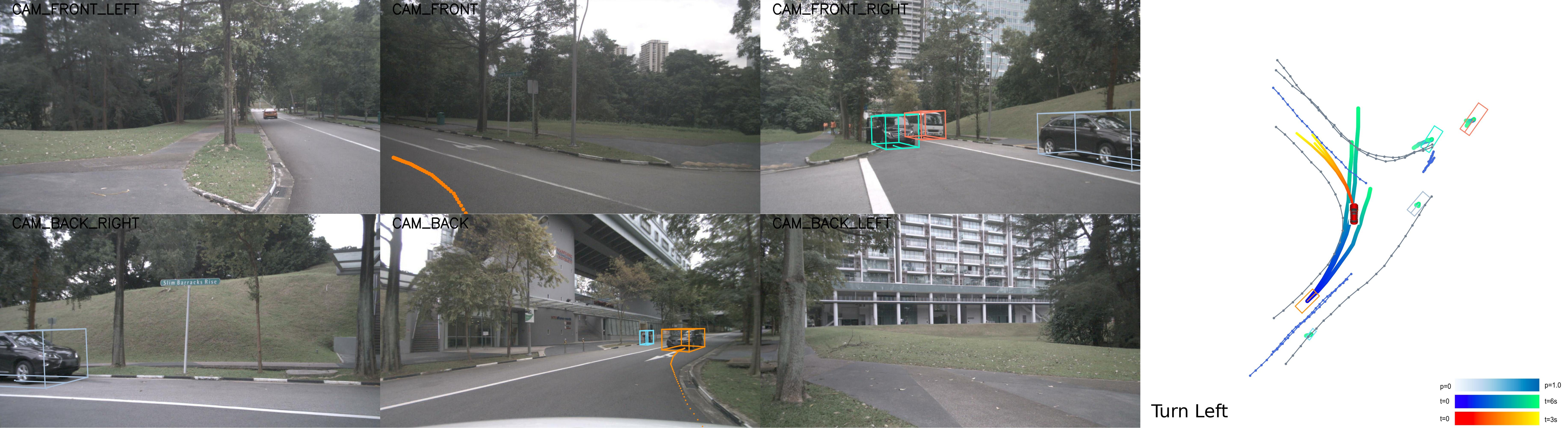}
  \end{subfigure}
  \begin{subfigure}{0.8\linewidth}
  \centering
  \includegraphics[width=1.0\linewidth]{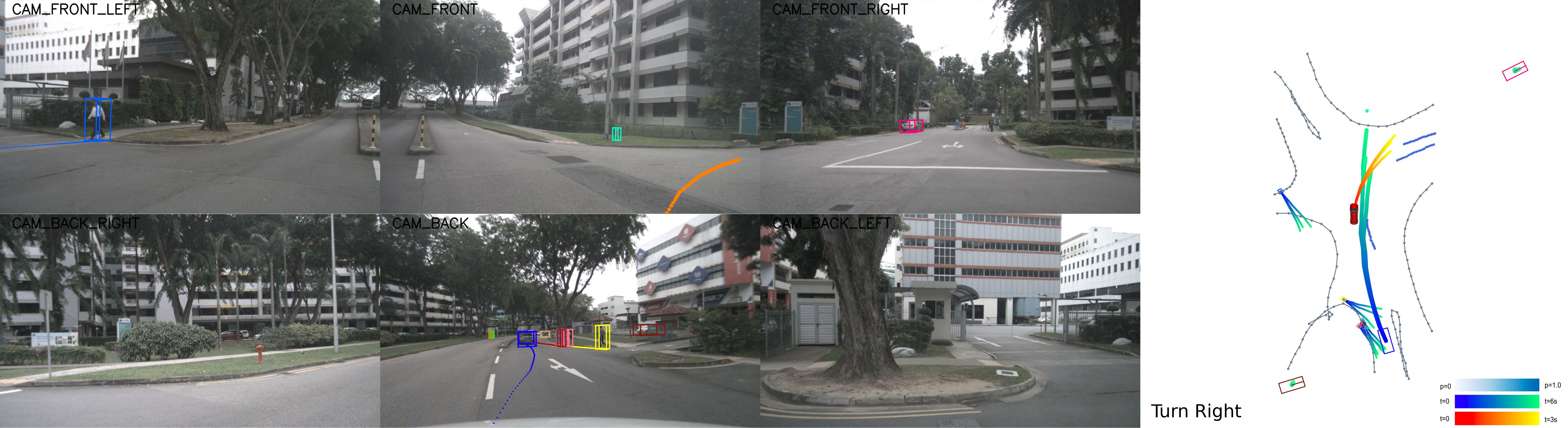}
  \end{subfigure}
  \begin{subfigure}{0.8\linewidth}
  \centering
  \includegraphics[width=1.0\linewidth]{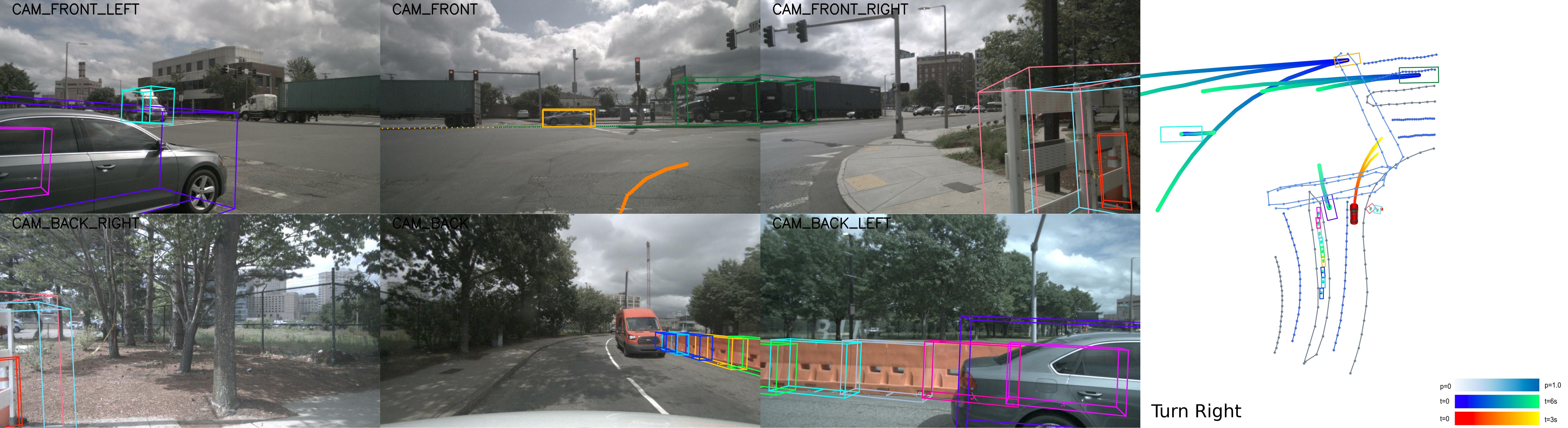}
  \end{subfigure}
  \begin{subfigure}{0.8\linewidth}
  \centering
  \includegraphics[width=1.0\linewidth]{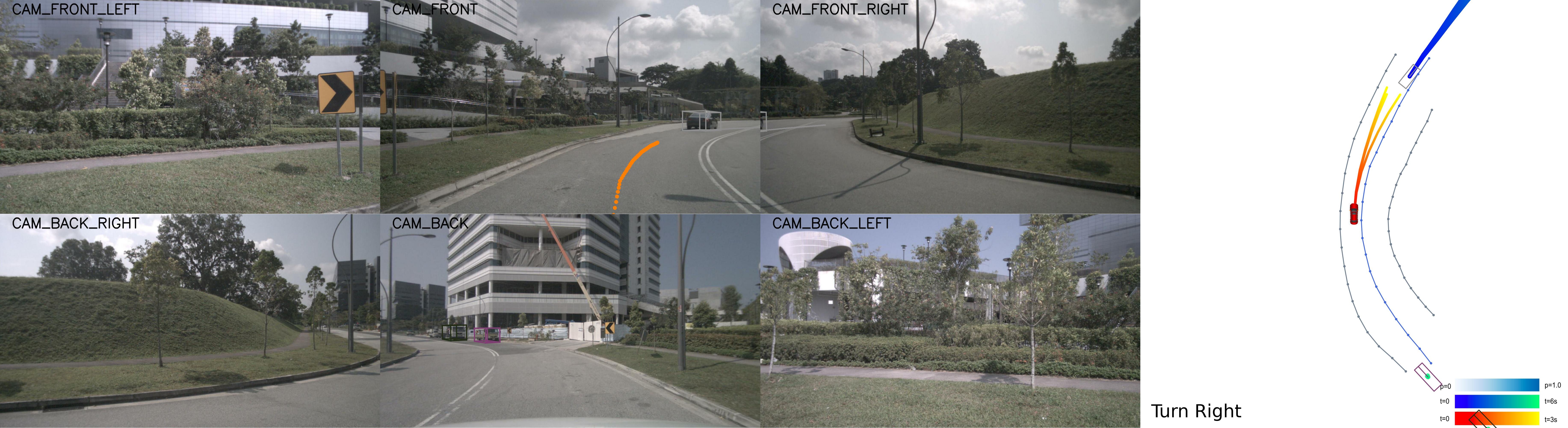}
  \end{subfigure}
  \caption{Visualization results. SparseDrive learnes different turning modes at intersections.}
  \label{fig:vis}
\end{figure}
\begin{figure}[htbp]
  \centering
  \begin{subfigure}{0.8\linewidth}
  \centering
  \includegraphics[width=1.0\linewidth]{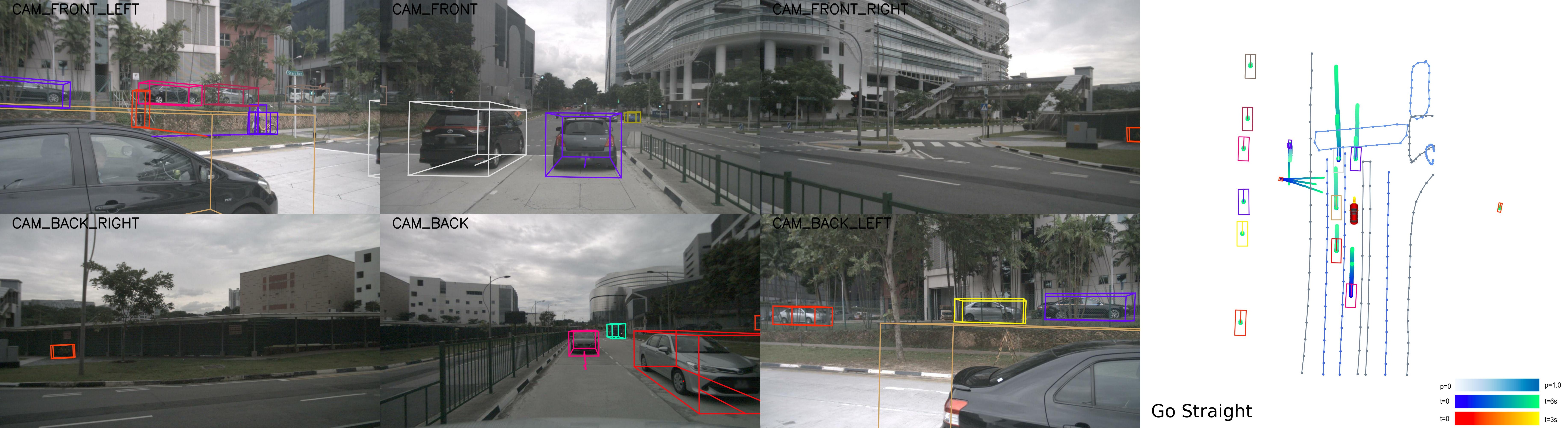}
  \end{subfigure}
  \begin{subfigure}{0.8\linewidth}
  \centering
  \includegraphics[width=1.0\linewidth]{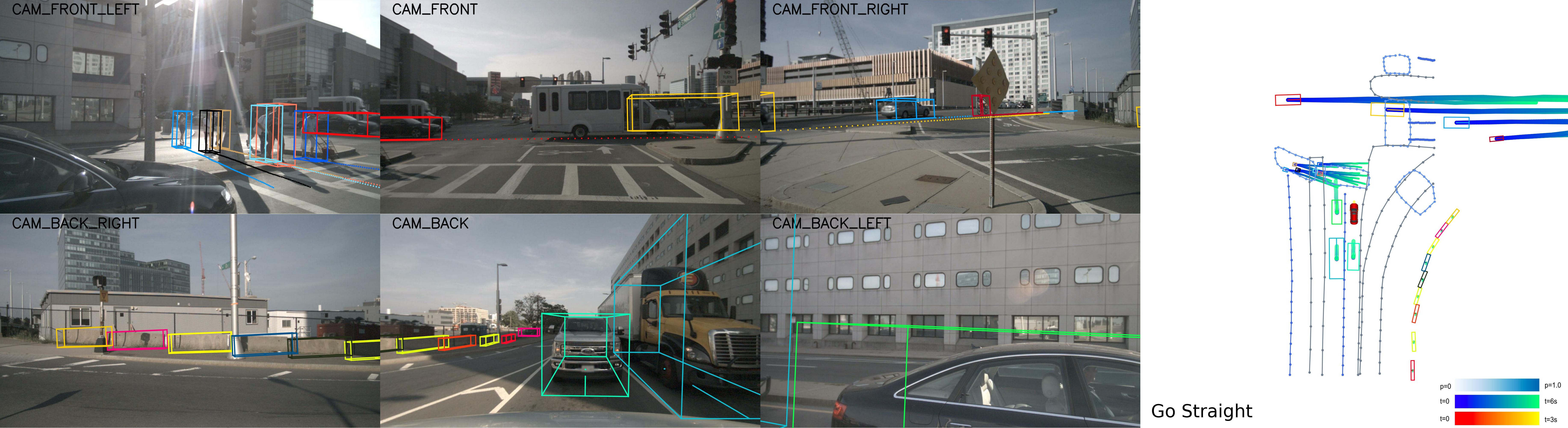}
  \end{subfigure}
  \begin{subfigure}{0.8\linewidth}
  \centering
  \includegraphics[width=1.0\linewidth]{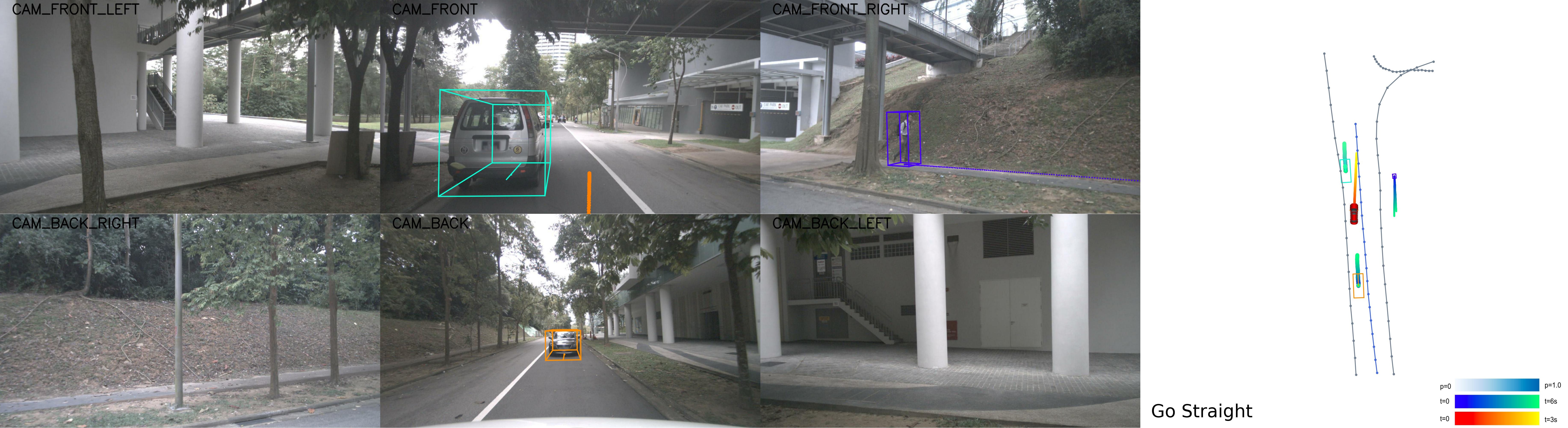}
  \end{subfigure}
  \begin{subfigure}{0.8\linewidth}
  \centering
  \includegraphics[width=1.0\linewidth]{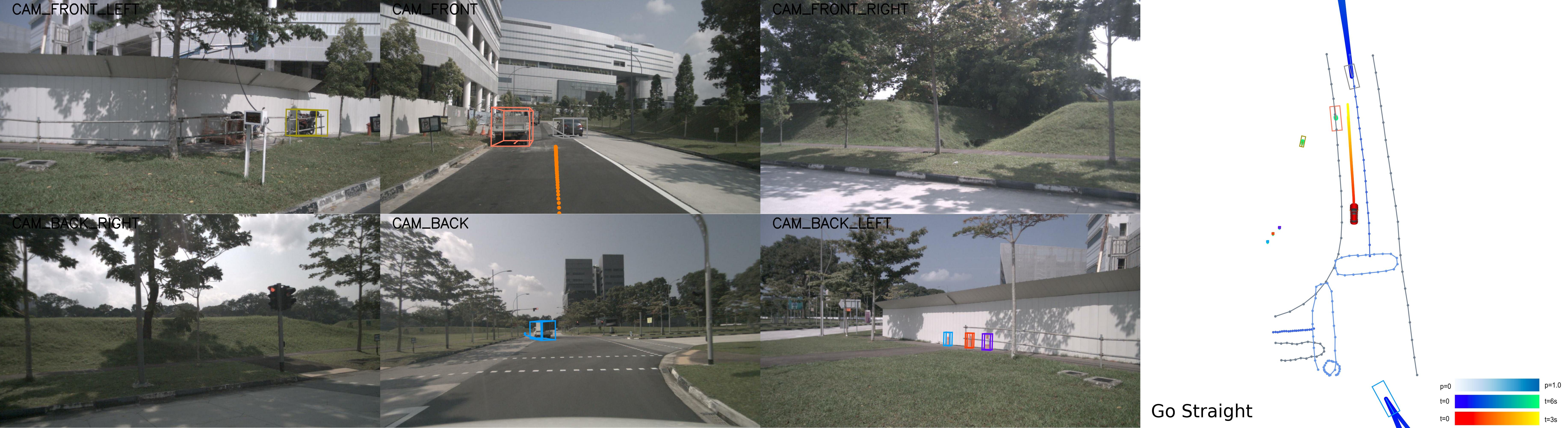}
  \end{subfigure}
  \caption{Visualization results. SparseDrive learns to yield to moving agents or avoid collision with obstacles.}
  \label{fig:vis}
\end{figure}
\end{document}